# Hierarchical and Decentralised Federated Learning

**Omer Rana, Theodoros Spyridopoulos**
Cardiff University, UK

**Nathaniel Hudson, Matt Baughman, Kyle Chard, Ian Foster**
University of Chicago, USA

**Aftab Khan**
Toshiba Research Europe

Federated Learning (FL) is a recent approach for distributed Machine Learning (ML) where data are never communicated to a central node. Instead an ML model (e.g., deep neural network) is initialised by a designated central (aggregation) node and shared with training nodes that have direct access to data of interest. These training nodes then perform small batches of training on their local data. Periodically, each training node submits ML model parameter/weight updates to the central node. The central node aggregates the parameters/weights to create a new global ML model that it then re-shares with the training nodes. This process can either take place indefinitely or be repeated until the ML model converges with respect to some evaluation metric (e.g., mean average error, accuracy).

As the Internet-of-Things (IoT) and other complex cyber-physical systems increase in scale, it is crucial that distributed learning approaches are also able to meet this increase in scale and diversity. FL is an attractive candidate solution because large, decentralised data are able to be kept private and maintained at the location where they are generated while still being used to train ML models. In addition, backhaul bandwidth requirements are reduced dramatically, because only the "learning"—i.e., parameters/weights—is transmitted, not the data upon which it is based. A more detailed background of FL can be found in McMahan and Ramage [8].

At a high level, different implementations of FL can be categorised as: (i) *cross-device* or *cross-silo* FL, and (ii) *horizontal* or *vertical* FL. Cross-device and cross-silo FL correspond to the hardware used as training nodes.

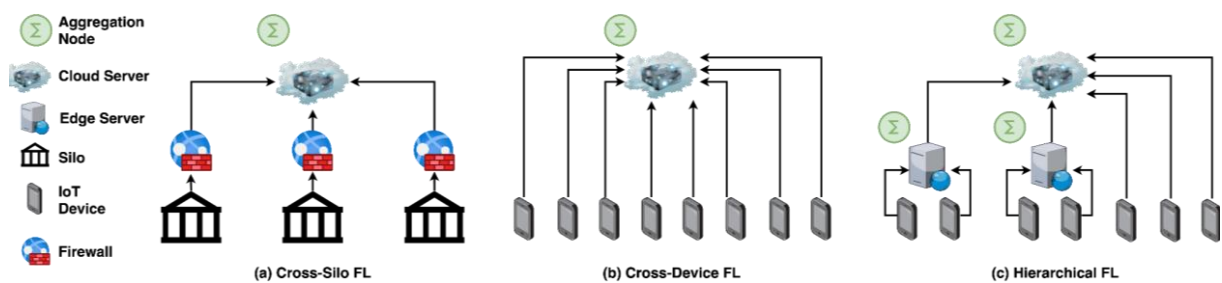

Fig 1: Overview of cross-silo, cross-device and hierarchical FL.

"Cross-device" FL generally considers many low-power, heterogeneous IoT devices (e.g., smart sensors, smartphones, wearables) that collect/maintain/host data for a single site. In this approach, FL is carried out across multiple devices, where each device independently trains the ML model ahead of aggregation. Cross-device FL is commonly used in edge and

IoT applications with the following key considerations: privacy of user data, sporadic availability of devices (i.e., some devices may be offline due to no power or poor connectivity), and heterogeneous hardware resources (i.e., low-powered edge devices). Conversely, "cross-silo" FL consists of several sites with ample compute resources. However, these centres (or silos) aim to train models from knowledge pooled across other centres without sharing raw, sensitive data. Unlike cross-device FL, the focus of cross-silo FL is privacy rather than resource constraints, since silos often represent large entities with sensitive/proprietary data and relatively stable compute resources. As an example, consider a hospital that uses ML models to assist physicians in detecting health conditions. The demographic distribution of clients seen by hospitals vary wildly based on socioeconomic conditions and geographic location. As such, one hospital may not be able to sufficiently train an ML model to predict for diverse clients if these ML models are trained on the hospital's own proprietary data. However, sharing raw health data is naturally sensitive. As such, applying aggregation techniques where several hospitals share ML model parameters can be a way to improve the robustness of the ML models owned by the hospitals without sharing raw data. Additional privacy-preserving techniques (e.g., homomorphic encryption) can be applied to further strengthen the privacy of the data. This is an example of cross-silo FL.

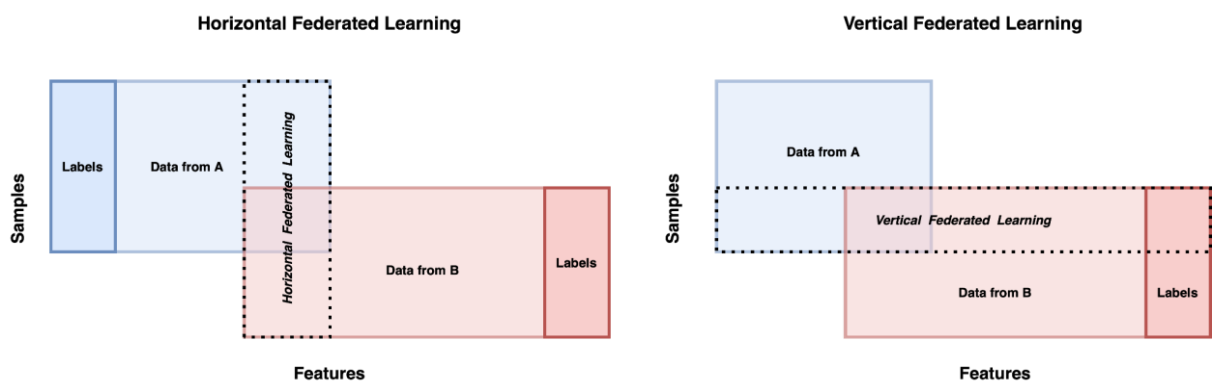

Fig 2: In horizontal FL there is a large overlap of features; In vertical FL there is a large overlap of sample IDs (users) across the two data sets.

The latter group of FL genres, horizontal FL and vertical FL, is centred on the distribution of data features in the FL process. Horizontal FL is the case where the data distribution among participating training nodes share the same feature space but vary in terms of which data samples belong on which training node. By contrast, vertical FL is where data on training nodes share the same sample space but vary in their feature spaces. An additional hybrid scenario is more common, wherein data on training nodes vary across both sample and feature spaces. For more information on horizontal, vertical, and hybrid FL, please refer to [17].

Hierarchical FL (H-FL) is a concept that can be thought of as essentially fusing the ideas of cross-device FL and cross-silo FL. Under H-FL, separate silos may perform their own intra-silo aggregations across their own low-power compute nodes that continually collect and train on data. However, the aggregated models from each of these silos may be further aggregated through inter-silo aggregations across silos without sharing of raw, sensitive data.

**The Aggregating Node**
In conventional FL processes, a node takes on the role of being the point of aggregation. This aggregation node is responsible for receiving ML model parameters, aggregating them, and sharing them with other training nodes. There are important challenges to consider for (i) the assignment of the aggregation node, and (ii) the design of the aggregation algorithm. For the former, the location and computational capacity of this aggregation node have implicit consequences for the performance of the entire FL process. For instance, if the communication channel between the aggregation node and training nodes is unreliable, then the aggregation node cannot reliably receive updated model parameters from the training nodes or share the global model. Further, the aggregation process can suffer from errors and noise injected into the model during transfer, errors introduced during training, and additional data modification caused by incorrectly configured devices or due to cyberattacks. Of course, many of these issues exist in centralized approaches as well.

Various aggregation algorithms [6] have been used to combine models in FL systems, across different geographically distributed devices. Approaches can vary from simply averaging the gradients of each of the independently-trained models (e.g., FedAvg – the approach proposed in Google's version of FL), to more complex weighted aggregation strategies taking account of system characteristics of devices and data characteristics across the distributed devices. Simple approaches for FL have their limitations. For instance, FedAvg has been shown to be biased towards contributing resources with data that is less representative of the total distribution. FedProx [20] on the other hand is created specifically to address variation in device hardware resources. FedProx augments the local training task with a proximal term in the local loss minimization objective. This ensures that the aggregation process makes use of parameters from devices that performed varying amounts of training based on their hardware resources with provable guarantees. Other variants of FL algorithms include Federated Stochastic Block Coordinate Descent (FedBCD) [13], which focuses on reducing the number of communication rounds by providing a tunable parameter that increases the number of local training rounds before the model communication between a device and the aggregation node takes place. When the data distribution is heterogeneous among client devices involved in training local models, it becomes necessary to "personalise" models to take account of this difference. In this context, the model can be split into local and global layers – referred to as adaptive "personalised" FL, using algorithms such as APFL [24], FedPer [6], and pFedMe [25].

Fairness also remains an important requirement in many FL systems, and represents: (i) equitable consideration of data from all sources; (ii) distribution of global models that ensures responsible use, i.e. where the global model can be scrutinised by all the nodes for fairness; (iii) tailoring of device-level training to suit the capabilities of each endpoint. Stochastic Agnostic Federated Learning (SAFL) [21] and the FedMGDA+ [22] algorithm have been proposed to address fairness during the FL training process. The SCAFFOLD algorithm has been proposed [23] to reduce the communication rounds required during model aggregation, using stateful variables in the distributed computing resources. Attention-augmented mechanisms are exploited in Attentive Federated Aggregation (FedAttOpt) to aggregate the knowledge generated from each computing resource based on the contribution made by each resource. Further discussion of these approaches can be found in Liu et al. [12].

When combining models, a number of different optimisation challenges are introduced. These include non-Independent and Identically Distributed (IID) data as well as differences in training data distributions between devices. The non-IID assumption introduces a number of challenges when aggregating models together, especially for the validity (and accuracy) of the aggregated model. For instance, with non-iid data distributions, a training device's local data will not contain a sufficient amount of balance across the data domain to train a model well enough on its own. For instance, if training an ML model to classify images with animals with the correct animal, a device with non-iid data may only have pictures of 1 or 2 animals. In such a case, it is crucial that training devices that participate in aggregation cover enough of the data set to produce a well-trained model upon aggregation.

A single aggregator node can become a bottleneck in many FL environments: (i) in terms of its performance, (ii) due to communication latency (and link failures) associated with sharing models between the aggregator and devices, and (iii) as a single point of failure. These bottlenecks become even more significant when considering an edge-computing network composed of heterogeneous devices (which may be mobile and thus intermittently connected) with differing computing capacities, spread across a large geographical area.

**Hierarchical FL**
Hierarchical FL (H-FL) consists of a master aggregator and a hierarchy of multiple aggregators to collectively combine trained local models, aiming to overcome constraints associated with a single aggregator node. The aggregation "tree" may be constructed based on the application context and constraints associated with computational capacity or device connectivity across independently managed aggregator nodes.

The location of aggregator nodes need not be pre-determined in an H-FL architecture. Based on characteristics of the communication network and the number (and types) of user requests being generated, aggregator nodes may be dynamically placed within the network to improve model accuracy and execution performance. Referred to as "worker aggregators" [7], we may differentiate between workers that: (i) are identical in their characteristics, i.e. worker nodes are identical in their resource requirements and can be hosted on any device in the network which has the resource capability to host the worker; (ii) are heterogeneous in their characteristics, e.g. they may require nodes which have specialist accelerator devices, e.g., to speed matrix-vector computation. The location of workers in (ii) can also be influenced by the types of models being aggregated (including the type of data on which these models are trained, e.g., image, audio, or numeric data). Such model uncertainty makes the placement of worker aggregators a challenging optimisation problem and may require dynamic updates to both the location of worker aggregators and the subsequent models being developed. An interesting approach is presented in [11], where worker placement is proposed as a hierarchical game, where the lower-level is an evolutionary game for resource allocation and the upper level is a Stackelberg game to determine a reward for players.

Placement of worker aggregators also assumes that the nodes which are selected to host them will be cooperative and support their hosting. This may be the case for a private implementation, but may not be the case in an open, commercial environment if the node owners are unwilling to handle the additional load (e.g., computation and communication) that such aggregation introduces. Incentive models—such as are in use for cryptocurrency mining—are likely to be required to ensure that such nodes can be rewarded in the aggregation process, which can vary from virtual credits that can be given to such nodes by a master aggregator, to the development of a market model that enables credits to be exchanged between nodes. These may be both monetary and non-monetary incentives, such as value alignment (e.g., SETI@home).

**Using resource-constrained devices**
In FL, devices used to train models at the edge of the network can be constrained in memory and processing capability. Further, centralised aggregation in large-scale edge networks can become inefficient due to the increased transmission latency. Training models under these constraints is in and of itself a difficult problem and may necessitate using less than the most performant models. This is especially problematic when using heterogeneous devices, as models must be tailored to work on the weakest computational device in the environment, or otherwise fairly and appropriately incorporate results from heterogeneous models. While methods exist to reduce computational and space requirements of models, these generally are focused on inference rather than model training or assume an already highly performant resource. Existing FL approaches typically assume that the aggregating node is centralised and is not limited by computational resource constraints.

Human experts may be used to assess quality of the developed model, especially where limited labelled data is available to carry out training. Further, human experts could support FL applications via active learning, where an online human-in-the-loop takes on the role of an oracle annotating periodically small subsets of unlabelled data. Active (continuous) learning has been long researched as a method to mitigate concept drift, a phenomenon where model performance deteriorates in time due to changes to the data and improve the overall model performance [18]. It is particularly effective in situations where the generation of labelled data is costly and time-consuming. Active learning-based FL has proven to provide comparable results to centralised approaches that use manually labelled data, requiring less labelling [19]. Understanding how an expert can work alongside an aggregator node provides an alternative to the more automated approach to hierarchical federation.

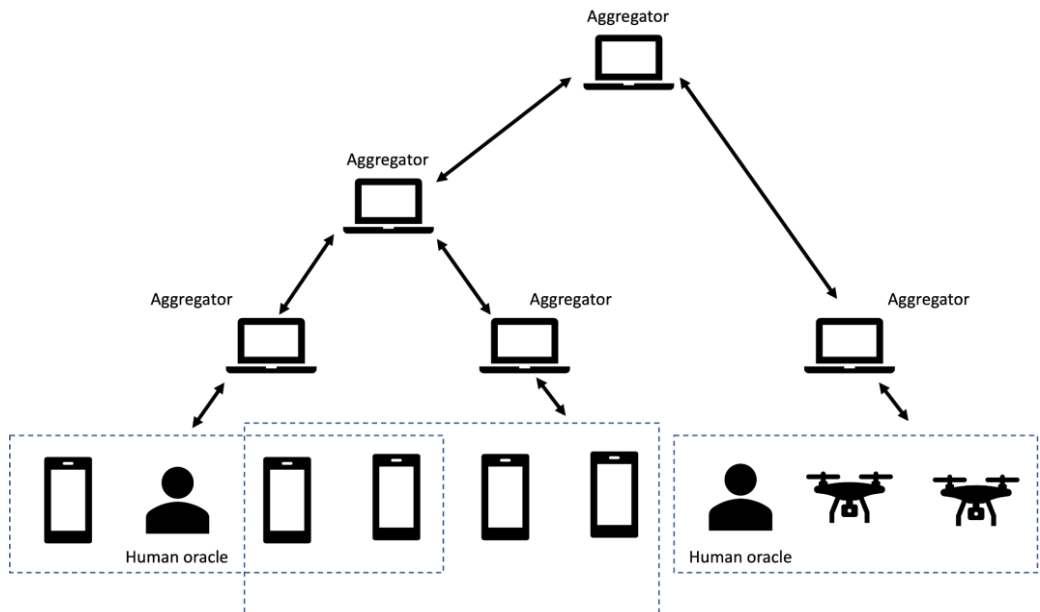

Fig 3: Active Learning-based Federated Learning with human oracle in the loop

**Software Frameworks for Hierarchical Federated Learning**
Over the past few years, several open-source-software frameworks have been developed to support research and deployment of FL processes. One of the first is Google's own TensorFlow Federated [4] which simulates FL processes on a single node. PaddleFL is an implementation of FL built on top of the PaddlePaddle deep learning framework [2]. PySyft is a framework put forward by the OpenMined group which focuses on privacy-preserving machine learning for FL (e.g., homomorphic encryption, differential privacy) [5]. FATE is an end-to-end FL software that supports FL network construction and visualisation [1]. These FL frameworks generally adopt conventional 2-tier, client-server architectures where there is one aggregation point for all the devices.

A recent contender, with potential for H-FL, in this area is FLoX (Federated Learning on FuncX), a federated Function-as-a-Service (FaaS) platform [10]. FLoX aims to be a highly modular framework with a single-call interface to make FL workflows accessible and usable for people with little experience with code, and ML experts alike. FLoX has been designed to execute on a serverless computing framework to better support diverse and distributed deployment environments. Serverless computing abstracts device heterogeneity and provides a high-level interface enabling computation (e.g., model training and aggregation) to be performed irrespective of the specific location at which it is executed. The funcX backend for FLoX has been demonstrated to scale from highly resource-constrained edge devices to exascale supercomputing centres. FLoX also allows users to easily specify the *amount of work* each endpoint should receive in accordance with those resources' capabilities. Additionally, due to the modularity provided by serverless environments, FLoX is capable of coordinating an FL workflow across any subset of resources and, in turn, enables users to aggregate multiple instantiations of FL to support hierarchical federated learning. FuncX provides functionality to launch multiple workers on each endpoint which allows for endpoint-level aggregation before global aggregation, as illustrated in Fig 4. This is a form of hierarchical learning that could be used to better utilise endpoint resources.

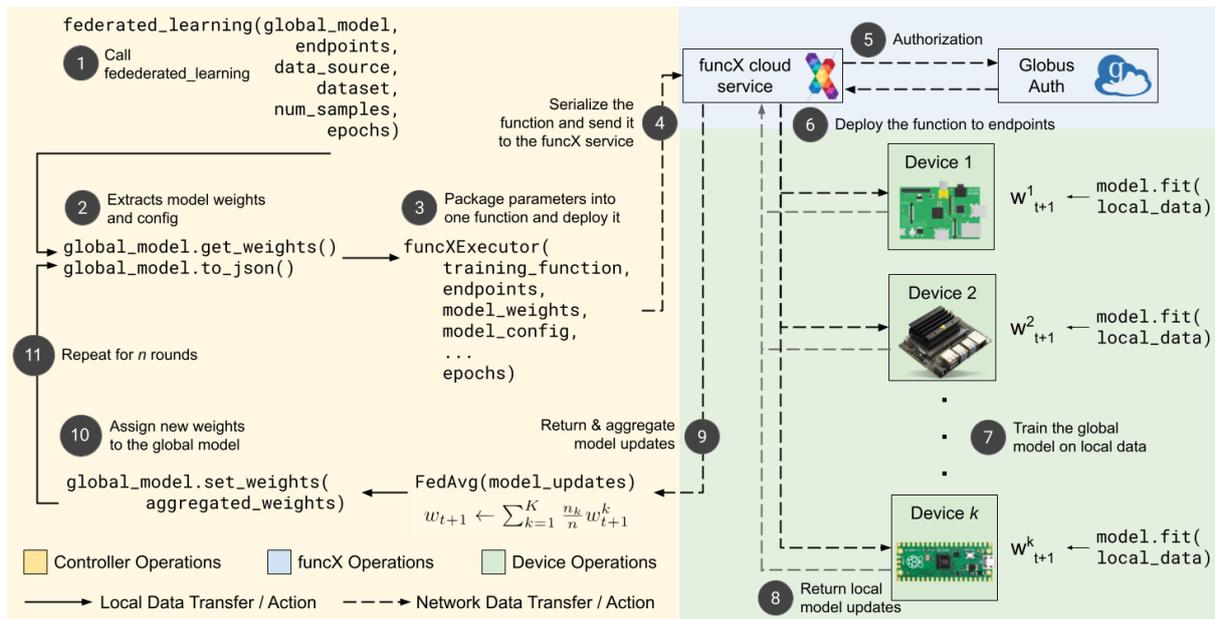

Fig 4: Software architecture for FLoX and its single function call interface [10].

**Applications**

Hierarchical FL is especially suited to applications executed in environments with complex and distributed cyber-infrastructure, made up of sensors and edge nodes. In these environments, sensors collect data to train models either directly at the sensor or by offloading to a nearby edge node with sufficient compute capacity. Within a particular region, there may be an edge node serving as an aggregation point specific to that region. Depending on the application, there may be benefit in aggregating across regions to improve the performance of the trained models. Relevant real-world use cases where hierarchical FL can be applied include (but are not limited to): (i) smart farming systems [14], (ii) smart traffic control systems [15], and (iii) smart energy grid systems [16].

Future smart farms will deploy sensors to collect real-time data regarding soil conditions and quality, moisture levels, temperature, pests, presence of weeds, etc. Edge nodes can then use this data to train models across a large-scale smart farm to control automated water irrigation systems, fertiliser dispensers, etc. Hierarchical FL can enable several smart farms to collaborate, such that models trained by separate smart farms can be combined to aggregate knowledge learned across smart farms.

Smart traffic-light control systems can be designed using FL. In such a system, traffic lights can learn optimal traffic-light control policies based on current traffic conditions. Federated Reinforcement Learning has been shown to improve the implicit trade-off between model performance and communication costs [15]. Hierarchical FL can enable cities to collaboratively train such FL-enabled infrastructure, with partial models being developed in a regional/ geographical context, and then aggregated to support particular application requirements. This may not always be possible and useful however, for instance where the models being aggregated represent very different types of environments. In this instance, an aggregated model may lead to a high error rate in practical use, and the nodes involved in undertaking aggregation may need to be limited.

Recently, hierarchical FL has been considered for the application of smart energy grids (namely, advanced metering infrastructure) for training models to disaggregate energy consumption signals into energy signals for individual appliances [14]. In this case, FL aggregation is considered at two different granularities: 2-tier and 3-tier aggregation. The 2-tier approach considers a Neighbourhood Area Network (NAN) aggregation node that is the sole aggregation point for participating smart homes. The 3-tier approach considers an additional layer of aggregation wherein aggregated models owned by the neighbourhood network nodes are collected and aggregated by another node in the system. Both approaches show comparable loss in accuracy of the model to a centrally trained model while greatly reducing the associated communication costs.

In addition to the aforementioned use cases, the UMBRELLA project[1] is an open, programmable smart-city sensing and wireless IoT testbed developed by Toshiba and supported by South Gloucestershire Council in the UK. Stretching from the University of West of England (UWE) Frenchay campus in the east to Bristol and Bath Science Park in the west, along the A4174 road corridor. UMBRELLA comprises over 230 edge nodes mounted on lamp posts. Each node is modular and consists of a processing module accompanied by different radio communication mechanisms (Bluetooth Low Energy, Wifi, LoRaWAN, and 4G) and multiple sensors (e.g., air quality sensors). In addition, over 100 of these nodes employ an additional processing module to train and aggregate models at the edge. The SYNERGIA[2] project builds on the UMBRELLA node to create a secure Edge IoT platform for multi-tenancy smart buildings. It extends the UMBRELLA platform's sensing capabilities, allowing the connection of battery-powered sensor devices (endpoints) to the node using its existing radios. The SYNERGIA system adopts a three-tier architecture (as illustrated in Fig 5). The three tiers are: the resource-constrained endpoint(s), the edge-computing layer with modest compute capacity and reliable network connectivity, and the backend tier with significant computing resources.

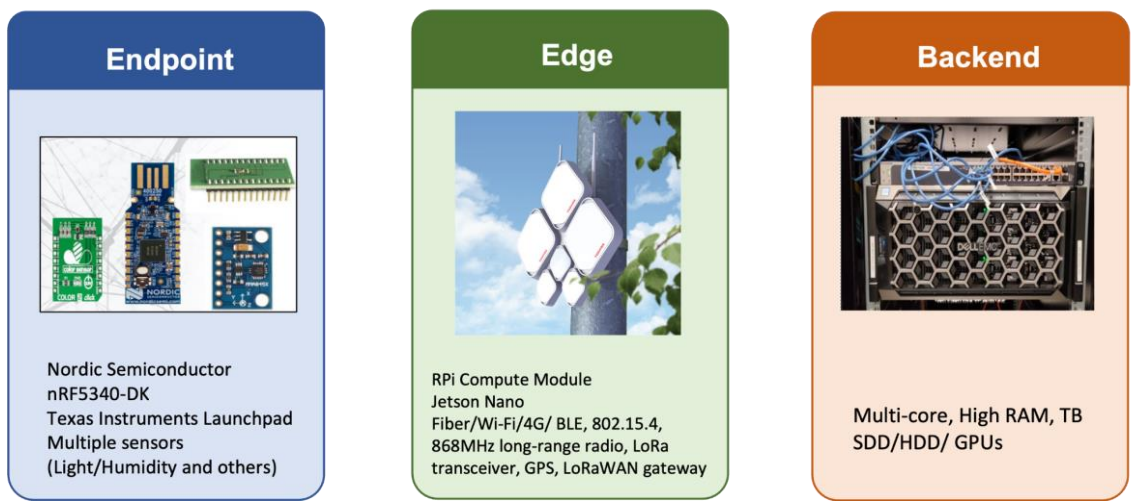

Fig 5: The three Tiers of the SYNERGIA platform

---

[1] https://www.umbrellaiot.com/
[2] https://synergia.blogs.bristol.ac.uk/

Within the SYNERGIA project, secure access is a key requirement for federating multiple resources in large-scale multi-tenancy environments. The project combines the establishment of secure device federation and AI mechanisms to identify *anomalous* device behaviour to support "trusted" federation. A FL approach is then followed to allow for knowledge sharing among the devices whilst preserving privacy and minimising data transfers in the network. In cases of low-confidence inferences on the nodes, the corresponding raw system data are collected centrally to the backend system for analysis by a human/expert-in-the-loop.

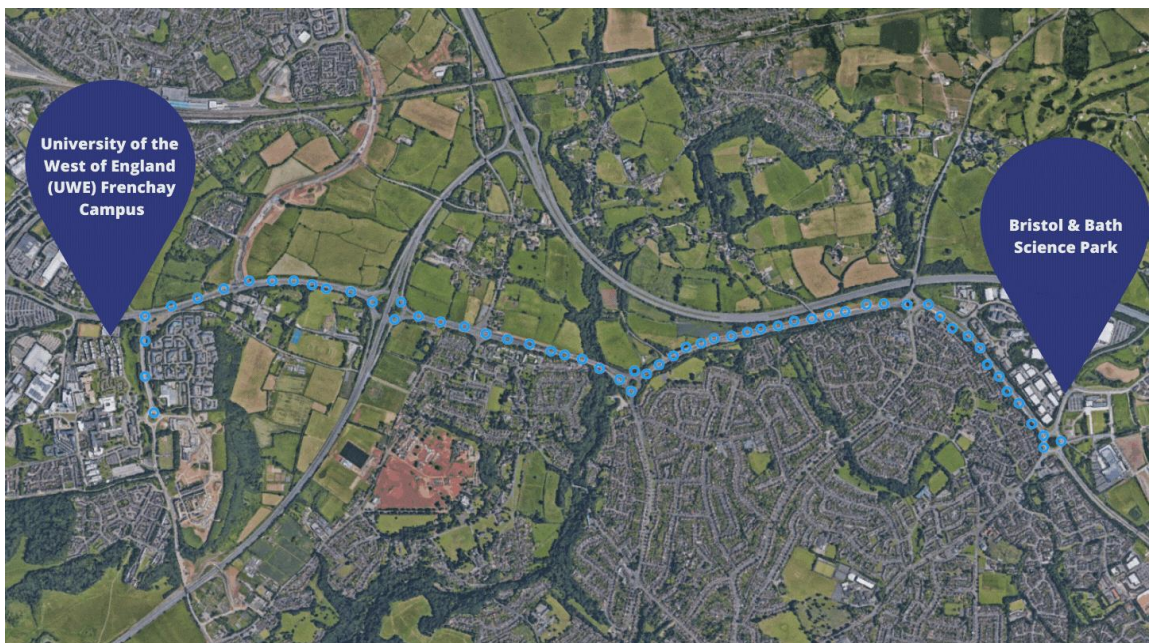

Fig 6: The UMBRELLA testbed: from UWE, Bristol Frenchay campus to the Bath Science Park in Bristol

**Conclusion**

Federated learning has shown enormous promise as a way of training ML models in distributed environments while reducing communication costs and protecting data privacy. However, the rise of complex cyber-physical systems, such as the Internet-of-Things, presents new challenges that are not met with traditional FL methods. Hierarchical Federated Learning extends the traditional FL process to enable more efficient model aggregation based on application needs or characteristics of the deployment environment (e.g., resource capabilities and/or network connectivity). It illustrates the benefits of balancing processing across the cloud-edge continuum.  Hierarchical Federated Learning is likely to be a key enabler for a wide range of applications, such as smart farming and smart energy management, as it can improve performance and reduce costs, whilst also enabling FL workflows to be deployed in environments that are not well-suited to traditional FL. Model aggregation algorithms, software frameworks, and infrastructures will need to be designed and implemented to make such solutions accessible to researchers and engineers across a growing set of domains.

H-FL also introduces a number of new challenges. For instance, there are implicit infrastructural challenges. There is also a trade-off between having generalised models and personalised models. If there exist geographical patterns for data (e.g., soil conditions in a smart farm likely are related to the geography of the region itself), then it is crucial that models used locally can consider their own locality in addition to a globally-learned model. H_FL will be crucial to future FL solutions as it can aggregate and distribute models at multiple levels to optimally serve the trade-off between locality dependence and global anomaly robustness.

## Acknowledgement
We are grateful to the Joe Weinman, Keith Jeffery & Lutz Schubert for their comments on an earlier version of this article.